# Multi-Camera Multi-Person Association using Transformer-Based Dense Pixel Correspondence Estimation and Detection-Based Masking

Daniel Kathein, Byron Hernandez, and Henry Medeiros
University of Florida, Gainesville, Florida

*Abstract*—Multi-camera Association (MCA) is the task of identifying objects and individuals across camera views and is an active research topic, given its numerous applications across robotics, surveillance, and agriculture. We investigate a novel multi-camera multi-target association algorithm based on dense pixel correspondence estimation with a Transformer-based architecture and underlying detection-based masking. After the algorithm generates a set of corresponding keypoints and their respective confidence levels between every pair of detections in the camera views are computed, an affinity matrix is determined containing the probabilities of matches between each pair. Finally, the Hungarian algorithm is applied to generate an optimal assignment matrix with all the predicted associations between the camera views. Our method is evaluated on the WILDTRACK Seven-Camera HD Dataset, a high-resolution dataset containing footage of walking pedestrians as well as precise annotations and camera calibrations. Our results conclude that the algorithm performs exceptionally well associating pedestrians on camera pairs that are positioned close to each other and observe the scene from similar perspectives. On camera pairs with orientations that are drastically different in distance or angle, there is still significant room for improvement.

## I. Introduction

Computer vision is an area of robotics and artificial intelligence that has been exponentially gaining traction. One sub-domain in particular, object detection, has become especially important with applications throughout video surveillance, robotics, self-driving cars, medical imaging, crop monitoring, and scene understanding [1]. Locating objects in images or videos, though relatively unchallenging for humans, can prove to be a difficult task for robots. Over the past few decades, deep learning has proven to be the most effective strategy for object recognition because computers can discover intricacies in significant volumes of image data and perform human-like analysis. Deep convolutional neural network-based technologies such as YOLO, SSD, and Faster R-CNN have been particularly successful in object recognition tasks [2]. These architectures can accurately identify people and objects within a single image and can be used within tracking-by-detection methods to analyze movement across multiple video frames. However, single-camera tracking is insufficient in many situations that cannot be captured by one camera. In order to perform tracking in these scenarios, a multi-camera association algorithm is necessary to associate people and objects with each other across multiple camera viewpoints.

Multi-camera tracking has widespread applications, especially in the field of security. Within airports, entertainment venues, schools, stadiums, and other locations, accurate surveillance systems can be installed to patrol a large area across several cameras and be helpful in crowd analysis and the early detection of potentially dangerous incidents. In such a vast space, one camera cannot sufficiently capture all the necessary data; thus, several cameras must work together and associate the visual information they capture. In airports, a high-accuracy multi-camera surveillance system could be applied at security checkpoints to accurately track passengers and baggage items to detect possible theft of items or abandoned bags. After reaching a certain threshold of confidence, the system could notify nearby security officers to investigate the situation and take action.

Not only does multi-view association give rise to advancements within automated security, but it also opens up avenues within other higher-level visual cognition tasks. The ability to determine whether two images are distinct perspectives of the same place is a fundamental cognitive ability that only the most advanced biological systems have, and for humans, it allows us to do the many crucial tasks in our daily lives, such as getting familiarized with new places. For robots, new multi-camera association technologies open up possibilities with visual reasoning tasks like scene construction, or the process of creating lifelike three-dimensional models of real-world objects and scenes [3].

Being able to look at two-dimensional images of a three-dimensional space and make inferences about what is seen is an invaluable skill in other realms as well, such as agriculture. High-accuracy multi-camera association technologies could be applied within aerial drones to efficiently perform crop mapping and yield estimation on large crop fields. It would also prove useful in high throughput plant phenotyping and disease detection, allowing farmers to improve their production productivity, achieve lower costs, and increase their output. When drones can optimally work together through multi-view association, each becomes more efficient and fewer drones are needed for the tasks at hand.

However, the task of matching people and objects in different views can be incredibly challenging for several reasons. Inherently, the complexity of multi-camera association depends primarily on the arrangement and number of cameras; however, in all scenarios, there is another universal issue: the same individual observed in two perspectives or time instants can look dramatically different. Existing machine learning architectures may interpret, for example, a single person wearing a multi-colored shirt as different people in multiple camera views. Conversely, systems may face the problem of inter-class confusion, where multiple people may be classified as the same person when observed in different camera views [4]. In addition to these issues, there are several other challenges for person re-identification and multi-view association systems: occlusion, illumination variance, pose variance, background clutter, person misalignment, and low camera resolution [5]. It is of the utmost importance that these issues be addressed within a high-accuracy multi-view association and tracking system.

Most current approaches tackling the task of determining which people and objects correspond to each other in multiple video views utilize homography-based methods to infer the transformation between them. A homography is a projective transformation that maps two planes with a three-by-three matrix [6], which can be used to project points from one image to another and find associations between individuals or items. However, these strategies fail to accurately represent a three-dimensional world in two dimensions, which is fundamentally nearly impossible using geometrical projections alone. A more robust solution is thus necessary. In this paper, we present an alternative approach involving dense correspondence estimation with neural networks and an underlying detection-based masking system to determine associations. This paper explores the extent to which it is possible to reproduce the human visual system's accuracy in the task of determining which people are associated with each other from multiple camera viewpoints. We hypothesize that our algorithm will be able to associate visual information collected with high accuracy for practical applicability.

## II. METHODS

This project explores a novel multi-camera association method and its application on a publicly available multi-camera dataset. The algorithm was developed on Python 3 using primarily the Kornia framework, which uses a Pytorch backbone. Development and testing were initially done locally on a Linux machine at the University of Florida, running Ubuntu 20.04 and containing the Nvidia RTX 3090 graphics card. NVIDIA Cuda version 11.6, NVCC version 10.1, and Torch version 1.8.2 were used on the computer.

### A. High Level Architecture

The proposed multi-view association algorithm uses dense correspondence estimation with a Transformer-based architecture and underlying detection-based masking to generate a set of correspondences between any pair of detections in the camera views provided. The keypoint matches and their respective confidence levels are used to determine an affinity matrix with the probabilities of matches between every detection pair. Finally, the Hungarian combinatorial optimization algorithm is applied to generate an assignment matrix with all the predicted associations between the camera views.

### B. Dense Correspondence Estimation

First, the algorithm uses dense correspondence estimation to detect keypoint matches between video frames in the camera pairs. Some well-known classical methods for this task of local feature matching include SIFT, SURF, BRIEF, and ORB [7]. These algorithms can detect hand-crafted features in computer vision tasks with reasonable accuracy. However, they often face issues when local features change significantly due to viewpoint or illumination variance. In recent years, deep learning has opened new horizons in local feature matching with the creation of algorithms like SuperGlue [8] and Local Feature Transformer [9]. Local Feature Transformer (LoFTR), a state-of-the-art detector-free feature matching algorithm built off Transformer neural networks, proves very useful in estimating dense correspondences between multiple images. The algorithm works at a coarse-to-fine level, first identifying pixel-wise dense matches between images and then refining the high-confidence matches at a sub-pixel level. This technique allows the method to outperform other methods of keypoint estimation to a great extent.

Through the transformer-based LoFTR architecture, the multi-view association algorithm first identifies the corresponding keypoints between video frames. Our algorithm uses the LoFTR's "outdoor" pre-trained model, which was trained on images taken outdoor and would be suitable for our data. The LoFTR configuration we used also includes a ResNetFPN backbone and four self and cross-attention layers. The LoFTR confidence threshold is left as a hyperparameter which is adjusted during algorithm testing. Before running LoFTR, masks are generated based on the homography mapping the transformation between the two camera views. In essence, this allows LoFTR to identify correspondences only between overlapping parts of the images. Once the video frames are converted to grayscale and fed into LoFTR, correspondences and their confidence levels are identified. Finally, masks are applied to detect the corresponding keypoints within only the bounding boxes of individuals in the video frames.

### C. Keypoint Confidence Refactoring

Next, the multi-view association pipeline includes keypoint confidence refactoring. The LoFTR correspondences and the homography mapping the transformation between the cameras are used for this task. First, a detected LoFTR keypoint from one camera is projected to a point on the other using the homography matrix. Then, the distance is calculated between these projected coordinates and the resulting LoFTR correspondence. A Gaussian function is used to generate new confidence values that are inversely proportional to the distance alone; the confidence level will be high when the distance is low, and vice versa. The LoFTR confidence and homography-based confidence values are then combined to create new refactored confidence values. A hyperparameter within the function is involved with this task and will be tweaked during testing.

*D. Affinity Matrix & Assignment Matrix*

Once the final confidence levels for the keypoint correspondences are calculated, the next task is determining the probability of a match between each detection pair in the two camera views. This involves generating an affinity matrix with several scalar values indicating the likelihood of matches between bounding box detections. Two primary metrics were developed to achieve this goal: an implementation of the Bayesian theorem (M4) and a unique Multi-frame (M5) approach.

Finally, given the affinity matrix, the Hungarian combinatorial optimization algorithm [10] is used to minimize the cost of determining associations between people in the camera views. It works by solving the underlying allocation problem and finding optimum pairwise correspondences in polynomial time. The final result is a list of associations between people and their bounding boxes in each video frame.

### III. DATASET

*A. Dataset Information*

We evaluated our multi-camera association algorithm on the WILDTRACK Seven-Camera HD Dataset, which includes footage of walking pedestrians in front of the main building at the ETH Zürich university in Switzerland [11]. This high-resolution dataset is composed of recorded video from seven stationary GoPro cameras with overlapping fields of view. Along with time-synchronized videos from each camera, WILDTRACK includes annotations for the 2D bounding boxes of pedestrians at a rate of two frames per second. It also includes high-precision intrinsic and extrinsic calibrations for each camera, making it ideal for assessing the performance of our multi-camera association algorithm [11]. Another benefit is that the dataset has radial distortion removed, so the cameras' curvature is effectively already corrected.

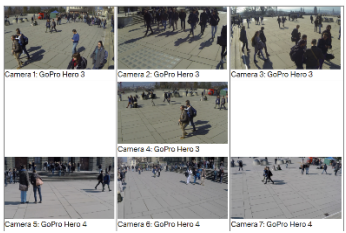
*Figure 1: WILDTRACK Camera Setup [11]*

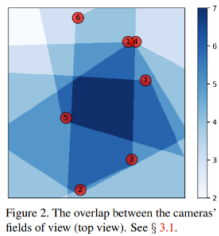
*Figure 2: WILDTRACK Camera Overlap [11]*

*B. Data Loader*

After obtaining a local copy of WILDTRACK, we designed and programmed a data loader into the algorithm to process the video frames, their annotations, and extrinsic/intrinsic calibration parameters. The frames were downscaled to 1280x720 resolution to improve testing efficiency and prevent memory errors. The annotations consist of the ground truth values for the bounding box detections, indicating the coordinates and size of the rectangles. The calibrations are crucial in generating homographies, which are used for masking and refactoring keypoint confidence levels within the algorithm. To generate homographies mapping the transformation between camera views, points are first generated in the 3D plane. WILDTRACK uses a 480x1440 grid with 2.5cm spacing and an origin at (-3.0, -9.0), so there are a total of 691,200 possible 3D points [11]. Rather than generating points on the ground plane to map the transformation of the floor, we generated points at Z = 40, which results in 100cm or around half the height of the average human. Using OpenCV's *projectPoints* function with these points and the intrinsic and extrinsic camera calibrations, we attain the projected 2D points in each individual camera. Then, given any camera pair, we used OpenCV's *findHomography* function with random sample consensus (RANSAC) to remove any outliers and calculate the homography. The RANSAC error tolerance threshold utilized was 10 pixels.

### IV. METHOD EVALUATION

*A. Experimental Design*

We assessed the performance of the multi-camera association algorithm on eight selected camera pairs in the WILDTRACK dataset, which have overlapping fields of view: 1 and 4, 1 and 6, 1 and 7, 2 and 3, 4 and 7, 5 and 6, 5 and 7, and 6 and 7. These were selected based on the amount of overlap between the video frames. During large-scale testing, we varied the algorithm hyperparameters for each camera pair to obtain the configuration with the best performance. The three primary hyperparameters tested were the LoFTR confidence threshold, refactor confidence distance coefficient, and the association metric. The LoFTR confidence threshold indicates the minimum confidence to keep corresponding keypoints generated by the LoFTR algorithm. The refactor confidence distance coefficient indicates the maximum allowed deviation of the homography-based point projection from the LoFTR correspondence. When this parameter's value is higher, the algorithm is more lenient if the homography-based point projection is further away. Finally, the association metric indicates how the affinity matrix should be generated: the Bayesian theorem (*M4*) or the Multi-frame (*M5*) approach. With 48 possible hyperparameter configurations tested on each camera pair, 384 total tests were completed to evaluate the algorithm on all eight pairs. Values tested in the experiment for each hyperparameter are given below:

*Table 1: Hyperparameter values tested for association on the WILDTRACK Seven-Camera HD Dataset*

| Hyperparameter | Values Tested |
|---|---|
| LoFTR Confidence Threshold | 0, 0.2, 0.4, 0.6 |
| Refactor Confidence Distance Coefficient | 0, 2, 5, 10, 20, 40 *0 = no refactoring* |
| Association Metric | M4, M5 |

*B. Evaluation Metrics*

In order to gauge the accuracy of the multi-view association algorithm and determine what hyperparameter configuration is most effective for each camera pair, we compute several performance evaluation metrics based on the association predictions and ground truth values: Precision, Recall, and F1-Score. These metrics are based on the number of true positives, false positives, and false negatives after association is completed for a camera pair. Precision indicates how precise the model is based on positive predictions and their actual values, whereas recall determines how many of the actual positives are labeled positive. Simply calculating the raw accuracy with the confusion matrix — correct predictions divided by total predictions — would not suffice due to the unbalanced dataset and abundance of true negatives in the data. This metric would misrepresent the algorithm's performance, hence the need to remove true negatives from our calculations.

## V. RESULTS

### A. Homography Results

Some of the homographies calculated from camera calibrations and the 3D point projections were very accurate, as depicted in *Figure 3*. Other homographies created warps that were only aligned with the ground plane; as shown in *Figure 4*, the tiles on the floor matched up between camera pairs but pedestrians were significantly distorted.

*Figure 3: Homography Warp for Camera Pair 1 and 4*

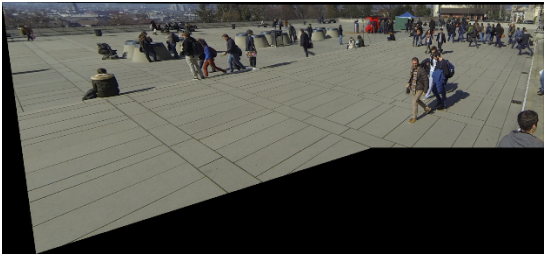

*Figure 4: Homography Warp for Camera Pair 1 and 7*

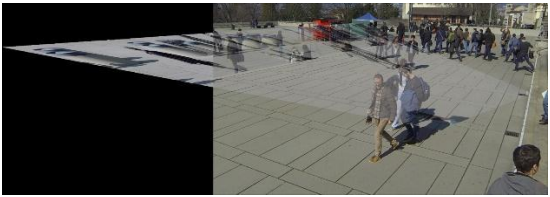

### B. Association Results

The results from the multi-camera association tests on the eight camera pairs are depicted below in Table 2.

*Table 2: Best Micro F-Score and other corresponding metrics for each camera pair and its configuration*

| Camera Pair | Precision (%) | Recall (%) | F-Score (%) | LoFTR Conf. Threshold | Refactor Conf. Distance Coefficient | Association Metric |
|---|---|---|---|---|---|---|
| 1, 4 | 98.72 | 97.05 | 97.88 | 0.4 | 10 | M5 |
| 1, 7 | 75.70 | 98.14 | 85.47 | 0 | 10 | M4 |
| 4, 7 | 42.79 | 94.33 | 58.87 | 0 | 10 | M5 |
| 1, 6 | 40.57 | 91.40 | 56.20 | 0 | 10 | M4 |
| 5, 7 | 25.03 | 94.23 | 39.56 | 0 | 20 | M5 |
| 6, 7 | 19.32 | 78.88 | 31.04 | 0 | 10 | M5 |
| 2, 3 | 17.22 | 93.96 | 29.10 | 0 | 10 | M4 |
| 5, 6 | 13.34 | 96.38 | 23.44 | 0 | 10 | M5 |

As depicted in *Figure 5* below, some camera pairs including 1 & 4 have near-perfect accuracy in associations. The algorithm handles even the occasional occlusion well, as shown with the three men who are still identified yet partially covering each other. However, other camera pairs, as depicted in *Figure 6*, do not have the same high accuracy; there are many missed associations and wrong associations.

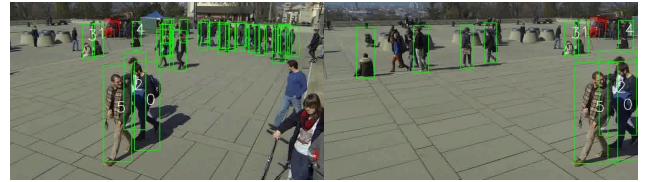

*Figure 5: Associations from Camera Pair 1 and 4 Frame (Best Pair)*

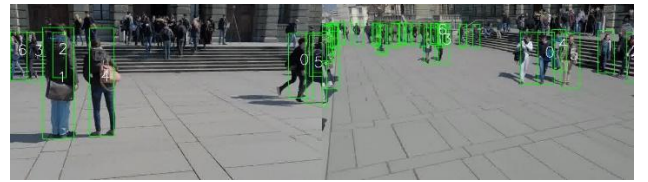

*Figure 6: Associations from Camera Pair 5 and 6 Frame (Worst Pair)*

## VI. DISCUSSION

Overall, the multi-view algorithm performed decently, with some camera pairs exhibiting significantly better performance than others. Camera 1 and 4, for example, achieved near-perfect performance as evidenced by the F1-Score metric of 97.88%. However, other pairs achieved F1-Score values below 50%, indicating that the algorithm was not performing well. This inconsistency could be attributed to the fact that the angle and distance between camera pairs are changing. For example, as shown in *Figure 5* and *Figure 6,* cameras 1 and 4 are positioned closer together and have less of an angle change than cameras 5 and 6. A large angle and distance between camera pairs could cause LoFTR to fail in generating accurate keypoint correspondences, as the Transformer-based architecture may simply not be able to perform local feature matching. Another possible reason why a drastic change in camera orientation may lower the performance of the algorithm is that it could affect the calculated homographies from camera calibrations. As depicted in *Figure 2*, as the scenes become more differently spaced and angled, the homography can distort the pedestrians when warping the video frames together. Even after calculating the homographies from projected 3D points up 100cm from the ground tiles — about half the height of a human — a point on a person's torso in one image would still be incorrectly projected using the homography to a point on the other image. This is a critical issue, as homographies are used within the algorithm to perform masking and keypoint confidence refactoring. However, because this is an inherent flaw with ground plane homographies, there may not be a simple solution given the structure of the dataset.

Another explanation behind the unsatisfactory performance of multi-view associations on a few of the tested camera pairs is the downscaling of the video frames to 1280x720 resolution. This change was made to increase efficiency and reduce memory errors on the machine; however, the downscaling would cause data to be lost and potentially cause the LoFTR algorithm to find less accurate correspondences.

Something to note is that ground truth bounding boxes and precise camera calibration data were provided within the dataset. When applying this multi-camera association algorithm to a real-time, real-world application, it would have to be paired with a powerful detector algorithm like YOLO, SSD, or Faster R-CNN to estimate bounding boxes [2]. Also, given that many camera setups do not have detailed camera calibrations, the homography-powered masking and keypoint refactoring may have to be reconsidered as they are essential parts of the multi-view association algorithm. One possible solution would be to use LoFTR correspondences to estimate a homography with RANSAC for this purpose. Then, without precise calibrations, the algorithm should still perform well.

Another factor to consider is that we are looking at each frame individually — temporal data is not considered in the algorithm. Thus, association IDs between frames can change, which is not ideal when we want consistent identities for bounding boxes throughout multi-camera tracking. Also, implementing temporal data for a new multi-camera tracking system based on the ideas discussed in this paper could potentially improve performance.

One path to explore in the future is modifying the algorithm's pairwise assignment method to determine associations from the affinity matrix. Rather than using the Hungarian optimization algorithm, a machine learning approach using a Deep Hungarian Network [12] or Deep Graph Matching Network [13] may prove beneficial in more accurately determining associations between pedestrians in the WILDTRACK camera pairs.

## VII. CONCLUSION

We were unable to reproduce the accuracy of the human visual system on all camera pairs within the WILDTRACK dataset. However, this method still seems very promising as it achieved exemplary performance on several camera pairs. We look forward to continued experimentation, algorithmic improvements, and assessments on other publicly available multi-camera datasets to determine if the method can be practically applicable.

REFERENCES

[1] A. R. Pathak, M. Pandey, and S. Rautaray, "Application of Deep Learning for Object Detection," *Procedia Computer Science*, vol. 132, pp. 1706–1717, 2018, doi: 10.1016/j.procs.2018.05.144.

[2] S. Srivastava, A. V. Divekar, C. Anilkumar, I. Naik, V. Kulkarni, and V. Pattabiraman, "Comparative analysis of deep learning image detection algorithms," *Journal of Big Data*, vol. 8, no. 1, May 2021, doi: 10.1186/s40537-021-00434-w.

[3] B. Mashburn and D.S. Blank, "Graphics + Robotics + AI = Fast, 3D Scene Construction," *Aaai.org*, 1999, Accessed: Jul. 27, 2022. [Online]. Available: https://www.aaai.org/Library/MAICS/1999/maics99-013.php

[4] S. Gong, M. Cristani, C. C. Loy, and T. M. Hospedales, "The Re-identification Challenge," *Person Re-Identification*, pp. 1–20, 2014, doi: 10.1007/978-1-4471-6296-4_1.

[5] A. Zahra, N. Perwaiz, M. Shahzad, and Fraz, Muhammad Moazam, "Person Re-identification: A Retrospective on Domain Specific Open Challenges and Future Trends," *arXiv.org*, 2022, doi: 10.48550/arXiv.2202.13121.

[6] X. Dai and S. Payandeh, "Tracked Object Association in Multi-camera Surveillance Network," *2013 IEEE International Conference on Systems, Man, and Cybernetics*, 2013, pp. 4248-4253, doi: 10.1109/SMC.2013.724.

[7] E. Karami, S. Prasad, and M. Shehata, "Image Matching Using SIFT, SURF, BRIEF and ORB: Performance Comparison for Distorted Images," *arXiv.org*, 2017, doi: 10.48550/arXiv.1710.02726.

[8] P.-E. Sarlin, D. DeTone, T. Malisiewicz, and A. Rabinovich, "SuperGlue: Learning Feature Matching with Graph Neural Networks," *arXiv.org*, 2019, doi: 10.48550/arXiv.1911.11763.

[9] J. Sun, Z. Shen, Y. Wang, H. Bao, and X. Zhou, "LoFTR: Detector-Free Local Feature Matching with Transformers," *arXiv.org*, 2021, doi: 10.48550/arXiv.2104.00680.

[10] H. W. Kuhn, "The Hungarian method for the assignment problem," *Naval Research Logistics Quarterly*, vol. 2, no. 1–2, pp. 83–97, Mar. 1955, doi: 10.1002/nav.3800020109.

[11] T. Chavdarova et al., "WILDTRACK: A Multi-camera HD Dataset for Dense Unscripted Pedestrian Detection," 2018 IEEE/CVF Conference on Computer Vision and Pattern Recognition, 2018, pp. 5030-5039, doi: 10.1109/CVPR.2018.00528.

[12] Y. Xu, A. Osep, Y. Ban, R. Horaud, L. Leal-Taixe, and X. Alameda-Pineda, "How To Train Your Deep Multi-Object Tracker," *arXiv.org*, 2019, doi: 10.48550/arXiv.1906.06618.

[13] R. Wang, J. Yan and X. Yang, "Learning Combinatorial Embedding Networks for Deep Graph Matching," *2019 IEEE/CVF International Conference on Computer Vision (ICCV)*, 2019, pp. 3056-3065, doi: 10.1109/ICCV.2019.00315.